%% file: main.tex
\pdfoutput=1

\documentclass[11pt]{article}

\usepackage[]{EACL2023}

\usepackage{times}
\usepackage{latexsym}

\usepackage[T1]{fontenc}

\usepackage[utf8]{inputenc}

\usepackage{microtype}

\usepackage{times}
\usepackage{latexsym}

\usepackage{booktabs}
\usepackage{adjustbox}
\usepackage{textcomp}
\usepackage{multirow}
\usepackage{xcolor}
\usepackage{colortbl}
\usepackage{todonotes}
\usepackage{enumitem}
\usepackage{footmisc}
\usepackage{amsmath}
\usepackage{comment}

\definecolor{ao(english)}{rgb}{0.0, 0.5, 0.0}
\definecolor{cornellred}{rgb}{0.7, 0.11, 0.11}
\definecolor{pf7}{RGB}{166, 118, 29}

\usepackage{tikz}
\usepackage{tikz-dependency}
\usetikzlibrary{%
  shapes,%
  arrows,%
  positioning,%
  calc,%
  automata%
}
\usepackage{expex}

\newcommand{\green}[1]{{\color{ao(english)}{#1} }}
\newcommand{\blue}[1]{{\color{blue}{#1} }}
\newcommand{\red}[1]{{\color{cornellred}{#1}}}
\newcommand{\cellgreen}[1]{{\cellcolor{green!25}{#1} }}
\newcommand{\furl}[1]{\footnote{\url{#1}}}

\newcommand{\parg}[1]{\ensuremath{\texttt{A#1}}}
\newcommand{\pargc}[1]{\ensuremath{\texttt{C-A#1}}}
\newcommand{\pargr}[1]{\ensuremath{\texttt{R-A#1}}}
\newcommand{\pargm}[1]{\ensuremath{\texttt{AM-#1}}}
\newcommand{\pargcstar}{{\texttt{C-X}}}
\newcommand{\pargrstar}{{\texttt{R-X}}}
\newcommand{\stag}[2]{\ensuremath{\left[\text{#1}\right]_\text{\tiny #2}}}

\newcommand{\shortprop}{PriMeSRL}
\title{PriMeSRL-Eval: A \textbf{Pr}act\textbf{i}cal Quality \textbf{Me}tric for \textbf{S}emantic \textbf{R}ole \textbf{L}abeling Systems \textbf{Eval}uation}

\author{First Author \\
  Affiliation / Address line 1 \\
  Affiliation / Address line 2 \\
  Affiliation / Address line 3 \\
  \texttt{email@domain} \\\And
  Second Author \\
  Affiliation / Address line 1 \\
  Affiliation / Address line 2 \\
  Affiliation / Address line 3 \\
  \texttt{email@domain} \\}
  
\author{Ishan Jindal\textsuperscript{a},  Alexandre Rademaker\textsuperscript{a}, Khoi-Nguyen Tran\textsuperscript{a},Huaiyu Zhu\textsuperscript{a}, \\ {\bf \large Hiroshi Kanayama\textsuperscript{a}, Marina Danilevsky\textsuperscript{a}, Yunyao Li\textsuperscript{b}{\thanks{\textsuperscript{b} Work done while at IBM Research}}} \\
\textsuperscript{a}IBM Research, 
\textsuperscript{b}Apple\\ 
  {\tt ishan.jindal@ibm.com, alexrad@br.ibm.com, kndtran@ibm.com,} \\{\tt huaiyu@us.ibm.com, hkana@jp.ibm.com, mdanile@us.ibm.com,} \\ {\tt yunyaoli@apple.com}}

\begin{document}
\maketitle

\begin{abstract}
Semantic role labeling (SRL) identifies the predicate-argument structure in a sentence. This task is usually accomplished in four steps: predicate identification, predicate sense disambiguation, argument identification, and argument classification. Errors introduced at one step propagate to later steps. Unfortunately, the existing SRL evaluation scripts do not consider the full effect of this error propagation aspect. They either evaluate arguments independent of predicate sense (CoNLL09) or do not evaluate predicate sense at all (CoNLL05), yielding an inaccurate SRL model performance on the argument classification task. In this paper we address key practical issues with existing evaluation scripts and propose a more strict SRL evaluation metric \textit{\shortprop}. We observe that by employing \shortprop, the quality evaluation of all SoTA SRL models drop significantly, and their relative rankings also change.  We also show that \shortprop successfully penalizes actual failures in SoTA SRL models.\footnote{\url{https://github.com/UniversalPropositions/PriMeSRL-Eval}}


\end{abstract}

\section{Introduction}\label{sec:intro}

Semantic Role Labeling (SRL) 
extracts predicate-argument structures from a sentence, where predicates represent relations (verbs, adjectives or nouns) and arguments are the spans attached to the predicate demonstrating ``who did what to whom, when, where and how.'' As one of the fundamental natural language processing (NLP) tasks, SRL has been shown to help a wide range of NLP downstream applications such as natural language inference \cite{zhang2020semantics,liu2022semantic}, question answering \cite{maqsud2014nerdle,yih2016value,zhang2020semantics,dryjanski-etal-2022-samsung}, machine translation \cite{shi2016knowledge,rapp:2022:LREC}, content moderation and verification \cite{calvo-figueras-etal-2022-semantics,fharook-etal-2022-hero}, information extraction \cite{niklaus-etal-2018-survey,DBLP:journals/corr/abs-2012-15243}.
In all of these applications, the quality of the underlying SRL models has a significant impact on the downstream tasks. Despite this, little study exists on how to properly evaluate the quality of practical SRL systems.

Given a sentence, a typical SRL system obtains predicate-argument structure by following a series of four steps: 1) predicate identification; 2) predicate sense disambiguation; 3) argument identification; and 4) argument classification. The predicate senses and its argument labels are taken from inventories of frame definitions such as Proposition Bank (PropBank) \cite{palmer2005proposition}, FrameNet~\cite{framenet}, and VerbNet~\cite{verbnet}. 

The correctness of SRL extraction is affected by the correctness of these steps. Consider the  example in Figure \ref{fig:srl-dep} using PropBank\footnote{In this paper we discuss SRL based on PropBank frames.} annotations:

\begin{figure}[tbhp]
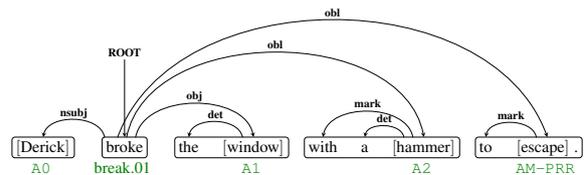

\centering
\resizebox{\linewidth}{!}{
\begin{dependency}[theme = simple, edge horizontal padding=1.5ex]
   \begin{deptext}[column sep=1em, row sep=.1ex]
         \stag{Derick}{} \&  broke \& the \& \stag{window}{}\& with\& a \&  \stag{hammer}{} \& to \& \stag{escape}{} . \\
         {\green{\parg{0}}} \& {\green{break.01}} \& {} \& {\green{\parg{1}}} \& {} \& {} \& {\green{\parg{2}}} \& {} \& {\green{\parg{M-PRR}}} \\
   \end{deptext}
   \depedge[edge start x offset=-6pt, label style={font=\bfseries,text=black}]{2}{1}{nsubj}{2ex}
   \depedge[label style={font=\bfseries,text=black}]{4}{3}{det}{2ex}
   \depedge[label style={font=\bfseries,text=black}]{2}{4}{obj}{2ex}
   \depedge[label style={font=\bfseries,text=black}]{7}{5}{mark}{2ex}
   \depedge[edge start x offset=-6pt, label style={font=\bfseries,text=black}]{7}{6}{det}{2ex}
   \depedge[edge start x offset=-6pt, label style={font=\bfseries,text=black}]{2}{7}{obl}{2ex}
   \depedge[label style={font=\bfseries,text=black}]{9}{8}{mark}{2ex}
   \depedge[edge start x offset=-12pt,arc angle=70, label style={font=\bfseries,text=black}]{2}{9}{obl}{2ex}
   \deproot[edge unit distance=3ex, label style={font=\bfseries,text=black}]{2}{ROOT}{2ex}
        \wordgroup{1}{1}{1}{a0}
        \wordgroup{1}{2}{2}{pred}
        \wordgroup{1}{3}{4}{a1}
        \wordgroup{1}{5}{7}{a2}
        \wordgroup{1}{8}{9}{amprr}
\end{dependency}}
\caption{An SRL example with head-based semantic roles on top of Universal Dependencies annotation.}
\label{fig:srl-dep}
\end{figure}


The SRL system must:
\begin{enumerate}
\itemsep-0.25em 
  \item Identify the verb `break' as a predicate
  \item Disambiguate its particular sense as `break.01',~\furl{https://verbs.colorado.edu/propbank/framesets-english-aliases/break.html} which has four associated arguments: \parg{0} (the breaker), \parg{1} (thing broken), \parg{2} (the instrument), \parg{3} (the number of pieces), and \parg{4} (from what \parg{1} is broken away).\footnote{Note that in PropBank each verb sense has a specific set of underspecified roles, given by numbers: \parg{0}, \parg{1}, \parg{2}, and so on. This is because of the well-known difficulty of defining a universal set of thematic roles \cite{slp}.}
  \item Identify each argument as it occurs (`Derick', `the window', etc.)
  \item Classify the arguments (`Derick' : \parg{0})
\end{enumerate}

Finally, this example has one additional modifier: the \pargm{PRP} (the purpose). Figure~\ref{fig:srl-dep} illustrate the same analysis on top of the universal dependencies annotations where only the tokens head of phrases are annotated with the proper argument. 

To obtain a completely correct predicate-argument structure both the predicate sense and all of its associated arguments need to be correctly extracted. Mistakes introduced at one step may propagate to later steps, leading to further errors. 

For instance, in the above example a wrong predicate sense `break.02' (\textit{break in or gain entry}) has not only a different meaning from `break.01' (\textit{break}), but also a different set of arguments. In many cases, even if an argument for a wrong predicate sense is labeled with the same numerical roles (\parg{1}, \parg{2}, etc), their meanings can be very different.  Therefore, in general, the labels for argument roles should be considered as incorrect when the predicate sense itself is incorrect.
However, existing SRL evaluation metrics (e.g.  \cite{hajic2009conll}) do not penalize argument labels in such cases.

The currently used evaluation metrics also do not evaluate \textit{discontinuous arguments} accurately. Some arguments in the PropBank original corpora have discontinuous spans that all refer to the same argument. This can happen for a number of reasons such as in verb-particle constructions. In a dependency-based analysis, these arguments ends up being attached to distinct syntactic heads \cite{surdeanu2008conll}. Take as an example the sentence, ``I know your answer will be that those people should be allowed to live where they please as long as they pay their full locational costs.'' For the predicate ``allow.01,'' the \parg{1} (action allowed) is the discontinuous span ``those people'' (\parg{1}) and ``to live where they please as long as they pay their full locational costs'' (\pargc{1}).  Existing evaluation metrics treat these as two independent labels. 

A similar problem exists for the evaluation of reference arguments (\pargrstar).  For example, in the sentence ``This is exactly a road that leads nowhere'', for the predicate ``lead.01'', the \parg{0} ``road'' is referenced by \pargc{0} ``that''.  If \parg{0} is not correctly identified, the reference \pargc{0} would be meaningless.

In this paper, we conduct a systematic analysis of the pros and cons of different evaluation metrics for SRL, including:
\begin{itemize}
    \item Proper evaluation of predicate sense disambiguation task;
    \item Argument label evaluation in conjunction with predicate sense;
    \item Proper evaluation for discontinuous arguments and reference arguments; and
    \item Unified evaluation of argument head and span.
\end{itemize}
We then propose a new metric for evaluating SRL systems in a more accurate and intuitive manner in Section \ref{sec:proposed}, and compare it with currently used methods in Section \ref{sec:comparison}.



\section{Existing Evaluation Metrics for SRL}\label{sec:metrics}
\input{existing-eval}

\section{The Proposed Approach}
\label{sec:proposed}



We propose \shortprop\ based on the following high-level rules to overcome the drawbacks in existing metrics:

\begin{enumerate}
 
  \setlength{\parskip}{0pt}
  \setlength{\itemsep}{0pt plus 1pt}
    \item Predicate senses are considered correct only when the predicate.sense is jointly correct instead of only the sense number is correct. (Table \ref{tab:pred_sense})
    \item Core arguments are considered as correct only when predicate sense is correctly identified. (Table \ref{tab:pred_sense}) 
    \item An argument of the form \pargcstar\ is considered together with its associated {\tt X} argument to cover the complete region of the argument. (Table \ref{tab:example_c})
    \item An arguments of the form \pargrstar\ is considered as a reference, so its correctness also depends on the correctness of the referenced {\tt X}. (Table \ref{tab:example_r})
\end{enumerate}

\input{psd_table}

\subsection{Predicate sense disambiguation evaluation}
Current evaluation metrics either do not evaluate the predicate sense disambiguation task (e.g. CoNLL05) or evaluate the sense number of the predicate (e.g. CoNLL09). In this section, we only contrast with the CoNLL09 evaluation script. 

First of all, what is predicate sense number? Similar to word senses in Wordnet \cite{fellbaum2010wordnet}, predicate senses in Propbank inside a frame file are generally ordered from most to least frequently used, with the most common sense numbered 01 \cite{pradhan2022propbank}. It means the sense numbers (e.g. 01, 02, 03, .. ) do not have associated semantic meaning. Sense number alone conveys the message that a particular meaning of the predicate is more common than others. Therefore, predicting/evaluating only the sense number alone does not make sense. However, it makes sense to a certain extent when predicate location is given and the task is only to disambiguate the sense of the predicate as proposed in the CoNLL09 shared task. But it has a consequence that it allows a sense number classifier to predict a sense number for a predicate that does not even exist in the associated frame file. Therefore, unknown sense number for a predicate does not have a semantic meaning, hence makes it unsuitable for practical use cases. For a practical end-to-end SRL system, it becomes necessary for the sense number classifier to predicate both the predicate location and associated sense number together (i.e. predicate.sense) so that the contextual meaning of the predicate is correctly captured, as performed in \cite{roth-lapata-2016-neural,li-etal-2018-unified,conia-etal-2021-unifying,conia2022probing}.  

The evaluation of the predicate sense disambiguation task of such practical systems using existing evaluation metrics is not optimal. To illustrate our point, we use the example in Figure \ref{fig:srl-dep}, where the gold predicate.sense is `break.01'. Suppose a practical SRL system predicts a predicate.sense label `pull.01'\footnote{It is unlikely for an SRL system to make such a diverse prediction, but we observe several cases in Table \ref{tab:psd_09_examples}.}, the existing CoNLL09 evaluation script gives a full score as the predicted sense number \textbf{01} exactly matches the gold sense number and fails to evaluate the semantic sense of the predicate. From the PropBank frame files, `break.01' means `break, cause to not be whole' and `pull.01' means `causing motion'. Despite this divergent meaning existing evaluation metric does not penalize. This is corrected by \shortprop\ by evaluating predicate.sense as a whole instead of evaluating sense number alone.  


\subsection{Argument evaluation with incorrect predicate sense}

Current metrics evaluate the arguments independent of the predicate sense. That is, they evaluate arguments as if the predicate location and sense are both correct. In practice, the predicates predicted by models can of course be wrong, and in such cases the corresponding core argument labels (\parg{0}, \parg{1}, etc.) generally do not refer to the correct argument - even if the label itself matches the gold label - and should be penalized. 
Contextual arguments, or adjunct arguments, such as \parg{M-LOC}, \parg{M-TMP}, etc, remain the same across different predicates and do not need to be penalized for predicate errors.  

Table~\ref{tab:pred_sense} illustrates the difference between \shortprop\ and existing evaluation metrics, CoNLL09 and CoNLL05.  For predicate sense evaluation, \shortprop\ is often equal to CoNLL09 (CoNLL05 does not measure this aspect.) \shortprop\ fixes a bug with CoNLL09 in the case where the lemma is wrongly identified (Example P3): The CoNLL09 script considers the label as correct as long as the ``predicate sense number'' is correct. 
For argument evaluation, both CoNLL09 head evaluation and CoNLL05 span evaluation wrongly mark all the arguments in examples P1, P2 and P3 as correct, despite the predicate sense being wrong. This is corrected by \shortprop.


\input{cstar_table}

\subsection{Evaluation of \pargcstar\ arguments}


An argument label with prefix \texttt{C-} is used in situations where an argument consists of multiple non-adjacent parts~\cite{surdeanu2008conll}.  If conceptually the whole argument should be labeled \texttt{X}, then operationally one part will get label \texttt{X} and the other parts get label \texttt{C-X}. The existing evaluations metrics treat all these labels as independent, which is incorrect as it increases the weight of these arguments and assigns partial credit when exact match is required. We now describe \shortprop\ for span-based and head-based evaluations.

\noindent\textbf{Span-based evaluation:} 
For an argument split into multipart spans with labels \texttt{X} and \texttt{C-X}, the complete span can be represented by the set of all tokens identified by these labels. The full set of tokens produced by the model should be compared to the set in the gold data, and a single credit should be assigned if these sets are equal.

\noindent\textbf{Head-based evaluation:} 
An argument with \texttt{X} and \texttt{C-X} parts has these as separate heads. A model prediction is considered correct if and only if all heads for this argument are correct, in which case it is given one whole credit. This evaluation does not distinguish between \texttt{X} and \texttt{C-X} and will penalize an argument if it has extra or missing parts.

Table~\ref{tab:example_c} compares \shortprop\ with ConNLL05 and ConNLL09 on seven examples. For span evaluation, the variances among labels with and without \texttt{C-} do not penalize the result, as long as the whole span is correct. For head evaluation, note that the denominators reflect the number of arguments rather than the number of split parts, and numerators count correct whole arguments.


\input{rstar_table}

\input{compare_table}

\begin{table*}[tbhp]
\centering
\begin{adjustbox}{width=0.91\linewidth}
\begin{tabular}{llll} \toprule
Id & Sentence  & Gold          & Predicted     \\ \midrule
1&\begin{tabular}[c]{@{}l@{}}He was able, now, to sit for hours in a chair in the living room and \blue{stare} \\ out at the bleak yard without moving.\end{tabular} & stare.01      & look.01       \\ \hline
2&\begin{tabular}[c]{@{}l@{}}She greeted her husband's colleagues with smiling \blue{politeness}, offering  \\  nothing.\end{tabular}                                & politeness.01 & minimalism.01 \\ \hline
3&\begin{tabular}[c]{@{}l@{}}It was a Negro section of \blue{peeling} row houses, store-front churches and\\  ragged children.\end{tabular}                          & peel.01       & peer.01       \\ \hline
4&He was calm, \blue{drugged}, and lazy.                                                                                                                             & drug.01       & dropper.01    \\ \hline
5&\begin{tabular}[c]{@{}l@{}}The walk and his fears had served to \blue{overheat} him and his sweaty armpits \\ cooled at the touch of the night air.\end{tabular}   & overheat.01   & soothe.01     \\ \hline
6&He did not resent their supervision or Virginia's sometimes \blue{tiring} sympathy.                                                                                & tire.01       & hiring.01    \\
\bottomrule
\end{tabular}
\end{adjustbox}
\caption{\citet{conia-etal-2021-unifying} model predictions on examples from CoNLL09 OOD set. All of these predicate senses are marked correct by the CoNLL09 evaluation script. \shortprop\ correctly penalizes all of these senses.}
\label{tab:psd_09_examples}
\end{table*}

 \input{compare_table_05}

\subsection{Evaluation of \pargrstar\ arguments}

An argument label with prefix \texttt{R-} indicates a reference argument; thus, \texttt{R-X} is a reference to the argument \texttt{X}.  For \texttt{R-X} to be correct, \texttt{X} must also be correct, but apart from this requirement, \shortprop\ treats them as separate arguments.


Table~\ref{tab:example_r} compares evaluating \pargrstar\ arguments using \shortprop\ with the metric used in CoNLL09 on 6 examples P1 through P6. For P1 in Table~\ref{tab:example_r} (Head Evaluation), CoNLL09 gives credit for correctly identified \pargr{0} for which no/incorrect \parg{0} is predicted, which is meaningless. Same is true for Span evaluation script CoNLL05. However, we do not penalize the correctly labeled main argument for incorrect \pargrstar.

\section{Comparisons with Existing Metrics}
\label{sec:comparison}
In this section, we discuss the effectiveness of existing SRL evaluation metrics and demonstrate how \shortprop\ differs in various use cases, using SoTA neural SRL models as test models.

\subsection{General settings}

For simplicity of comparison with existing results, we assume the gold predicate location is given for all the experiments following  \citet{shi2019simple, jindal2020improved, conia2022probing}. However, \shortprop\ is able to handle missing or spurious predicates. We use \citet{conia-etal-2021-unifying, blloshmi2021generating, jindal-EtAl:2022:LREC} as SoTA SRL models. 

\subsection{Datasets}

We show the impact of evaluating with \shortprop\ on the CoNLL09 and CoNLL05 datasets. Table~\ref{tab:dataset} shows the percentage of \pargcstar\ and \pargrstar\ arguments in each split of the different datasets. Note that these arguments make up $<3\%$ of the total arguments; $5.09\%$ total of the arguments in CoNLL05 test, and $2.95\%$ in CoNLL09 test. Therefore, we expect to observe an F1 drop of at most about $3$ and $5$ points on the argument classification subtask due to mishandling  \pargcstar\ and \pargrstar\ arguments for CoNLL09 and CoNLL05 datasets, respectively.

\subsection{Evaluation} 
\noindent\textbf{Predicate sense disambiguation} PSD column in Table~\ref{tab:p_09} compares the impact of \shortprop\ w.r.t to the existing evaluation script on EN subset of CoNLL09 dataset using SoTA SRL models. As expected, we observe a quality drop in predicate sense disambiguation (PSD) both for in-domain and out-of-domain (OOD) sets. Surprisingly, on OOD set, we observe a significant quality drop of an average of $\sim~5$ F1 points for all the SRL models that significantly lowers the SoTA performance on OOD set. This shows that the existing SRL models sill have a lot of room for improvement. 

Further, In Table~\ref{tab:psd_09_examples} we show some example instances from CoNLL09 dataset, which have correct sense numbers (01) but wrong predicate.sense. They are marked as correct by the CoNLL09 evaluation script. 
For example, the first row shows how the difference between  "stare.01" (\textit{looking intently~\furl{https://verbs.colorado.edu/propbank/framesets-english-aliases/stare.html}}) and "look.01" (\textit{causal look~\furl{https://verbs.colorado.edu/propbank/framesets-english-aliases/look.html}}) is ignored. 
While these two at least share the same underlying meaning (\textit{look}), in row~5 the model's prediction of `soothe.01' means the opposite of the gold label `overheat.01' -- yet the existing CoNLL09 evaluation script still marks this as correct. It is therefore clear that predicate sense should be evaluated using  predicate.sense, instead of sense number alone. 

\begin{table*}[]
\centering
\begin{adjustbox}{width=\linewidth}
\begin{tabular}{@{}llllll@{}}
\toprule
Input Sentence                                                                                                                & SRL model prediction                                                                                                                        & \begin{tabular}[c]{@{}l@{}}Existing \\ eval score\end{tabular} & \shortprop\ & Application                                                                             & Output                                                 \\ \midrule
\multirow{4}{*}{\begin{tabular}[c]{@{}l@{}}{[}S1{]} XYZ company  \\ bought \$2.4 billion in \\ Fannie Mae bonds.\end{tabular}} & \begin{tabular}[c]{@{}l@{}}{[}XYZ company{]}\parg{0} \\ {[}bought{]}\red{sell.01}\\ {[}\$2.4 billion in Fannie Mae bonds{]}\parg{1}\end{tabular}      & 3/3                                                            & 0/3     & \begin{tabular}[c]{@{}l@{}}QA\\ Who bought \\ Fannie Mae bonds?\end{tabular}            & None                                                   \\ \cmidrule(l){2-6} 
                                                                                                                              & \begin{tabular}[c]{@{}l@{}}{[}XYZ company{]}\parg{0} \\ {[}bought{]}\red{buy\_out.03} \\ {[}\$2.4 billion in Fannie Mae bonds{]}\parg{1}\end{tabular}      & 2/3                                                            & 0/3     & \begin{tabular}[c]{@{}l@{}}QA\\ Who bought \\ Fannie Mae bonds completely?\end{tabular} & \begin{tabular}[c]{@{}l@{}}XYZ \\ company\end{tabular} \\ \midrule
\begin{tabular}[c]{@{}l@{}}{[}S2{]} XYZ company \\ bought out \$2.4 billion\\ in Fannie Mae bonds.\end{tabular}              & \begin{tabular}[c]{@{}l@{}}{[}XYZ company{]}\parg{0} \\ {[}bought{]}{buy\_out.03}  out \\ {[}\$2.4 billion in Fannie Mae bonds{]}\parg{1}\end{tabular} & 3/3                                                            & 3/3     & \multirow{5}{*}{\begin{tabular}[c]{@{}l@{}}NLI\\ Does S2 entails S3?\end{tabular}}      & \multirow{5}{*}{Yes}                                   \\ \cmidrule(r){1-4}
\begin{tabular}[c]{@{}l@{}}{[}S3{]} XYZ company\\ bought \$2.4 billion in \\ Fannie Mae bonds.\end{tabular}                   & \begin{tabular}[c]{@{}l@{}}{[}XYZ company{]}\parg{0} \\ {[}bought{]}\red{buy\_out.03} \\ {[}\$2.4 billion in   Fannie Mae bonds{]}\parg{1}\end{tabular}    & 2/3                                                            & 0/3     &                                                                                         &                                                        \\ \bottomrule 
\end{tabular}
\end{adjustbox}
\caption{Example illustrations of impact of SRL errors on downstream applications captured by the new evaluation method, where \red{Red color} represents the wrong prediction by SRL model; QA: Question Answering; NLI: Natural Language Inference. All outputs are incorrect.}
\label{tab:application}
\end{table*}

\noindent\textbf{Argument head evaluation} Argument classification column in Table~\ref{tab:p_09} compares the impact of \shortprop\ w.r.t to the existing evaluation script on EN subset of CoNLL09 dataset using SoTA SRL models.  We observe a quality drop in argument classification task both for in-domain and out-of-domain (OOD) sets with a significant quality drop of an average of $\sim~8$ F1 points on OOD set. This drop on OOD argument classification task is expected because part of this error is propagated from the predicate sense disambiguation task that itself is significant.  It is interesting to note that although the major contribution of argument classification drop is due to error propagating from earlier stage but there is consistent drop of $\sim~1.5$ and $\sim~3$ F1 points because of penalizing correct arguments with wrong predicate sense for in-domain and OOD sets, respectively. 

Since the performance drop is not uniform we observe a change in the relative ranking of the SRL models. As an example, with CoNLL09 evaluation script SRL models \citet{blloshmi2021generating} and \citet{jindal-EtAl:2022:LREC} get similar scores ($~83.0$ F1) on OOD set whereas \shortprop\ clearly shows the difference in performance. Further, with \shortprop\ it is clear that the quality of existing SRL systems are not as good as previously thought, especially on OOD data, and there is still much room for improvement.

\noindent\textbf{Argument span evaluation} Similarly, we compare the impact of \shortprop\ w.r.t to the existing evaluation script on the CoNLL05 dataset using SoTA SRL models in Table~\ref{tab:p_05}. Since CoNLL05 does not evaluate predicate sense, we do not observe the impact of incorrect PSD on argument classification. Therefore, the only drop of argument classification is due to incorrect handling of \pargcstar\ and \pargrstar\ arguments. Although Table~\ref{tab:dataset} shows that the total number of \pargcstar\ and \pargrstar\ in the CoNLL05 dataset is $\sim 5\%$ of the total number of arguments, we only observe a slight drop in quality evaluation ($<1\%$) with \shortprop. This is because on argument span evaluation, \shortprop\ is similar to CoNLL05 (except in a few cases as described in last row-block of Tables~\ref{tab:example_c} and \ref{tab:example_r}.) Similar to the comparison with the CoNLL09 dataset, we again observe a change in the relative ranking of the SRL models.



\subsection{Discussion}

The existing evaluation metrics for SRL are disconnected from the actual practical performance of the SRL models. Therefore, makes it difficult to choose a better quality SRL model for the downstream application. Current evaluation metrics do not pay sufficient attention to the error propagation aspect of four-staged SRL task, instead evaluate the steps independently and linearly combine them to compute the overall SRL system score. Our analysis in Tables \ref{tab:p_09} and \ref{tab:p_05} reveals that one should not mistaken the linear combination of the performance of individual steps for the actually overall quality of the SRL system.  

This does not mean that the existing evaluation metric are not useful, instead these metrices provide an individual evaluation of each step that itself is quite important to improve the quality of individual steps and hence the quality of the SRL system. However, we argue that for a real-world NLP system that utilize SRL as one of its components, it is important to understand the quality of semantic roles in relation with their predicate sense disambiguation. Table \ref{tab:application} illustrate the impact of such SRL errors on two downstream applications (question answering and natural language inference). Existing evaluation scripts overlook these SRL errors and assign high score despite the predicted predicate-argument structure is meaningless and lead to incorrect choice of SRL model for downstream application.

\section{Conclusion}

In this paper we highlighted key issues with existing SRL evaluation metrics and showed that our proposed method, \shortprop, scores SoTA SRL models in a more accurate and intuitive manner. By releasing our evaluation code, we plan to promote these metrics in the community in order to improve the evaluation quality for SRL systems that contribute to downstream applications.

\section{Limitations}


We have shown the impact of our proposed new evaluation metrics in the current SoTA SRL models ranking. To further validate the impact this work, we plan to conduct an in-depth study on how downstream applications' performance relates to the evaluation metrics in future work. 

We acknowledge that the problems we have pointed out for previous evaluation metrics are not bugs, but rather design decisions given the timing of the shared tasks and the limitations on datasets and methods. Consider, for instance, that a unified syntactic dependency annotation schema like Universal Dependencies \cite{nivre-etal-2016-universal} was unavailable before October 2014. Given that, and the space limitation in this paper, we didn't present a deep discussion on the impact of UD compared to previously used syntactic dependencies schemas.

\bibliography{anthology,custom}
\bibliographystyle{acl_natbib}

\appendix

\input{appendix}



\end{document}

%% file: existing-eval.tex

Most of the existing evaluation metrics came from shared tasks for the development of systems capable of extracting predicates and arguments from natural language sentences. In this section, we summarize the approaches to SRL evaluation in the shared tasks from SemEval and CoNLL

\subsection{Senseval and SemEval}

SemEval (Semantic Evaluation) is a series of evaluations of computational semantic analysis systems that evolved from the Senseval (word sense evaluation) series. 

\textbf{SENSEVAL-3} \cite{litkowski-2004-senseval} addressed the task of automatic labeling of semantic roles and was designed to encourage research into and use of the FrameNet dataset. 
The system would receive as input a target word and its frame, and was required to identify and label the frame elements (arguments). The evaluation metric counted the number of arguments correctly identified (complete match of span) and labeled, but did not penalize those spuriously identified.  An overlap score was generated as the average of proportion of partial matches.

\textbf{SemEval-2007} contained three tasks that evaluate SRL. 
Task 17 and 18 identified arguments for given predicates using two different role label sets: PropBank and VerbNet \cite{pradhan-etal-2007-semeval}. 
They used the \verb|srl-eval.pl| script from the CoNLL-2005 scoring package \cite{carreras-marquez-2005-introduction} (see below).  
Task 19 consists of recognizing words and phrases that evoke semantic frames from FrameNet and their semantic dependents, which are usually, but not always, their syntactic dependents. The evaluation measured precision and recall for frames and frame elements, with partial credit for incorrect but closely related frames. Two types of evaluation were carried out. The first is the label matching evaluation. The participant's labeled data were compared directly with the gold standard labeled using the same evaluation procedure used in the previous SRL tasks at SemEval. The second is the semantic dependency evaluation, in which both the gold standard and the submitted data were first converted to semantic dependency graphs and compared. 

\textbf{SemEval-2012} \cite{kordjamshidi-etal-2012-semeval} and \textbf{SemEval-2013} \cite{kolomiyets-etal-2013-semeval} introduced the `Spatial Role Labeling' task, but this is somewhat different from the standard SRL task and will not be discussed in this paper. Since \textbf{SemEval-2014} \cite{marelli-etal-2014-semeval}, a deeper semantic representation of sentences in a single graph-based structure via semantic parsing has superseded  the previous `shallow' SRL tasks.

\subsection{CoNLL}

The \textbf{CoNLL-2004} shared task \cite{carreras-marquez-2004-introduction} was based on the PropBank corpus, comprising six sections of the Wall Street Journal part of the Penn Treebank \cite{kingsbury-palmer-2002-treebank} enriched with predicate–argument structures. 
The task was to identify and label the arguments of each marked verb. The precision, recall and F1 of arguments were evaluated using the \verb|srl-eval.pl| program. For an argument to be correctly recognized, the words spanning the argument as well as its semantic role have to be correct. The verb argument is the lexicalization of the predicate of the proposition. Most of the time, the verb corresponds to the target verb of the proposition, which is provided as input, and only in few cases the verb participant spans more words than the target verb. This situation makes the verb easy to identify and, since there is one verb with each proposition, evaluating its recognition overestimates the overall performance of a system. For this reason, the verb argument is excluded from evaluation. The shared task proceedings does not details how non-continuous arguments are evaluated. In \textbf{CoNLL-2005} \cite{carreras2005introduction} a system had to recognize and label the arguments of each target verb. The evaluation method remained the same as CoNLL-2004, using the same evaluation code. 

The \textbf{CoNLL 2008} shared task \cite{surdeanu-etal-2008-conll} was dedicated to the joint parsing of syntactic and semantic dependencies. The shared task was divided into three subtasks: (i) parsing of syntactic dependencies, (ii) identification and disambiguation of semantic predicates, and (iii) identification of arguments and assignment of semantic roles for each predicate. SRL was performed and evaluated using a dependency-based representation for both syntactic and semantic dependencies. 

The official evaluation measures consist of three different scores: (i) syntactic dependencies are scored using the labeled attachment score (LAS),
(ii) semantic dependencies are evaluated using a labeled F1 score, and (iii) the overall task is scored with a macro average of the two previous scores. The semantic propositions are evaluated by converting them to semantic dependencies, i.e., a semantic dependency from every predicate
to all its individual arguments were created. These dependencies are labeled with the labels of the corresponding arguments. Additionally, a semantic dependency from each predicate to a virtual
ROOT node was created. The latter dependencies are labeled with the predicate senses. This approach guarantees that the semantic dependency structure conceptually forms a single-rooted, connected (not necessarily acyclic) graph. More importantly, this scoring strategy implies that if a system assigns the incorrect predicate sense, it still receives some points for the arguments correctly assigned. Several additional evaluation measures were applied to further analyze the performance of the participating systems. The \emph{Exact Match} reports the percentage of sentences that are completely correct, i.e., all the generated syntactic dependencies are correct and all the semantic propositions are present and correct. The \emph{Perfect Proposition F1} score entire semantic frames or propositions. The ratio between labeled F1 score for semantic dependencies and the LAS for syntactic dependencies.

As in CoNLL-2008, the CoNLL-2009 shared task \cite{hajic-etal-2009-conll} combined syntactic dependency parsing and the task of identifying and labeling semantic arguments of verbs or nouns for six more languages in addition to the original English from CoNLL-2008. Predicate disambiguation was still part of the task, whereas the identification of argument-bearing words was not. This decision was made to compensate for the significant differences between languages and between the annotation schemes used. The evaluation of SRL was done similar to CoNLL-2008.

%% file: psd_table.tex
\begin{table}[]
\centering
\begin{adjustbox}{width=\linewidth}
\begin{tabular}{|ccclllll|}
\multicolumn{8}{l}{\#   text = Yesterday, John bought a car.} 
\\ \hline
\multicolumn{1}{|c|}{\multirow{2}{*}{ID}} & \multicolumn{1}{c|}{\multirow{2}{*}{FORM}} & \multicolumn{1}{c|}{\multirow{2}{*}{FLAG}}                                                    & \multicolumn{1}{c|}{\multirow{2}{*}{\begin{tabular}[c]{@{}c@{}}PRED\\ SENSE\end{tabular}}} & \multicolumn{4}{c|}{Predicate-argument prediction}                         \\ \cline{5-8} 
\multicolumn{1}{|c|}{}                    & \multicolumn{1}{c|}{}                       & \multicolumn{1}{c|}{}                                                                         & \multicolumn{1}{c|}{}                                                                      & 
\multicolumn{1}{l|}{Gold} & \multicolumn{1}{l|}{P1} & \multicolumn{1}{l|}{P2} & P3 \\ \hline
\multicolumn{1}{|l|}{1}                   & \multicolumn{1}{l|}{Yesterday}              & \multicolumn{1}{l|}{\_}                                                                       & \multicolumn{1}{l|}{\_}                                                                    & \multicolumn{1}{l|}{\texttt{TMP}} & \multicolumn{1}{l|}{\texttt{TMP}}      & \multicolumn{1}{l|}{\texttt{TMP}} & \texttt{TMP} \\ \hline
\multicolumn{1}{|l|}{2}                   & \multicolumn{1}{l|}{,}                      & \multicolumn{1}{l|}{\_}                                                                       & \multicolumn{1}{l|}{\_}                                                                    & \multicolumn{1}{l|}{\_}     & \multicolumn{1}{l|}{\_}          & \multicolumn{1}{l|}{\_}    & \_  \\ \hline
\multicolumn{1}{|l|}{3}                   & \multicolumn{1}{l|}{John}                   & \multicolumn{1}{l|}{\_}                                                                       & \multicolumn{1}{l|}{\_}                                                                    & \multicolumn{1}{l|}{\parg{0}}     & \multicolumn{1}{l|}{\parg{0}}          & \multicolumn{1}{l|}{\parg{0}}  & \parg{0}     \\ \hline
\multicolumn{1}{|l|}{4}                   & \multicolumn{1}{l|}{bought}                 & \multicolumn{1}{l|}{Y}                                                                        & \multicolumn{1}{l|}{buy.01}                                                                & 
\multicolumn{1}{l|}{buy.01} & \multicolumn{1}{l|}{\red{\textit{buy\_out.03}}} & \multicolumn{1}{l|}{\red{\textit{buy.05}}} & \red{\textit{sell.01}} \\ \hline
\multicolumn{1}{|l|}{5}                   & \multicolumn{1}{l|}{a}                      & \multicolumn{1}{l|}{\_}                                                                       & \multicolumn{1}{l|}{\_}                                                                    & \multicolumn{1}{l|}{\_}     & \multicolumn{1}{l|}{\_}          & \multicolumn{1}{l|}{\_}   & \_   \\ \hline
\multicolumn{1}{|l|}{6}                   & \multicolumn{1}{l|}{car}                    & \multicolumn{1}{l|}{\_}                                                                       & \multicolumn{1}{l|}{\_}                                                                    & \multicolumn{1}{l|}{\parg{1}}     & \multicolumn{1}{l|}{\parg{1}}          & \multicolumn{1}{l|}{\parg{1}}    & \parg{1}  \\ \hline
\multicolumn{1}{|l|}{7}                   & \multicolumn{1}{l|}{.}                      & \multicolumn{1}{l|}{\_}                                                                       & \multicolumn{1}{l|}{\_}                                                                    & \multicolumn{1}{l|}{\_}     & \multicolumn{1}{l|}{\_}          & \multicolumn{1}{l|}{\_}   & \_  \\ \hline
\multicolumn{1}{c}{}                     & \multicolumn{7}{c}{}                                                                                                                                                                                                                                                                                              \\ 
\cline{2-8} 
\multicolumn{1}{c|}{}                     & \multicolumn{1}{c|}{\multirow{7}{*}{\begin{tabular}[c]{@{}c@{}}Predicate \\ Evaluation\end{tabular}}} & 
\multicolumn{1}{c|}{\multirow{3}{*}{R}}                                             
    & \multicolumn{1}{l|}{CoNLL05}                                                                     & \multicolumn{4}{c|}{do not evaluate}                                                   \\ \cline{4-8}
\multicolumn{1}{c|}{}                     & \multicolumn{1}{c|}{}                       & \multicolumn{1}{c|}{}                                                                         & \multicolumn{1}{l|}{CoNLL09}                                                                     & \multicolumn{1}{l|}{1/1}    & \multicolumn{1}{l|}{0/1}      & \multicolumn{1}{l|}{0/1}     & 1/1      \\ \cline{4-8} 
\multicolumn{1}{c|}{}                     & \multicolumn{1}{c|}{}                       & \multicolumn{1}{c|}{}                                                                         & \multicolumn{1}{l|}{\shortprop}                                                                     & \multicolumn{1}{l|}{1/1}    & \multicolumn{1}{l|}{0/1}    & \multicolumn{1}{l|}{0/1}       & \cellgreen{0/1}      \\ \cline{3-8}
\multicolumn{1}{c|}{}        & \multicolumn{1}{c|}{}        & \multicolumn{6}{c|}{} \\ \cline{3-8}
\multicolumn{1}{c|}{}                     & \multicolumn{1}{c|}{}                       & \multicolumn{1}{c|}{\multirow{3}{*}{P}}                                                 & \multicolumn{1}{l|}{CoNLL05}                                                                     & \multicolumn{4}{c|}{do not evaluate}                   \\ \cline{4-8}
\multicolumn{1}{c|}{}                     & \multicolumn{1}{c|}{}                       & \multicolumn{1}{c|}{} &                                                                    \multicolumn{1}{l|}{CoNLL09}                                                                     & \multicolumn{1}{l|}{1/1}    & \multicolumn{1}{l|}{0/1}   & \multicolumn{1}{l|}{0/1}       & 1/1      \\ \cline{4-8} 
\multicolumn{1}{c|}{}                     & \multicolumn{1}{c|}{}                       & \multicolumn{1}{c|}{}                                                & \multicolumn{1}{l|}{\shortprop}                                                                     & \multicolumn{1}{l|}{1/1}    & \multicolumn{1}{l|}{0/1}    & \multicolumn{1}{l|}{0/1}       & \cellgreen{0/1}      \\ \cline{2-8} 
\multicolumn{8}{l}{}\\
\cline{2-8} 
\multicolumn{1}{c|}{}                     & \multicolumn{1}{c|}{\multirow{7}{*}{\begin{tabular}[c]{@{}c@{}}Argument \\ Evaluation\end{tabular}}} & \multicolumn{1}{c|}{\multirow{3}{*}{R}}                                                 & \multicolumn{1}{l|}{CoNLL05}                                                                     & \multicolumn{1}{l|}{3/3}    & \multicolumn{1}{l|}{3/3}   & \multicolumn{1}{l|}{3/3} & 0/3                                                   \\ \cline{4-8}
\multicolumn{1}{c|}{}                     & \multicolumn{1}{c|}{}                       & \multicolumn{1}{c|}{}                                                                         & \multicolumn{1}{l|}{CoNLL09} &                                                                     \multicolumn{1}{l|}{3/3}    & \multicolumn{1}{l|}{3/3}  & \multicolumn{1}{l|}{3/3}        & 3/3      \\ \cline{4-8} 
\multicolumn{1}{c|}{}                     & \multicolumn{1}{c|}{}                       & \multicolumn{1}{c|}{}                                                                         & \multicolumn{1}{l|}{\shortprop}                                                                     & \multicolumn{1}{l|}{3/3}    & \multicolumn{1}{l|}{\cellgreen{1/3}}   & \multicolumn{1}{l|}{\cellgreen{1/3}}      & \cellgreen{1/3}      \\ \cline{3-8}
\multicolumn{1}{c|}{}        & \multicolumn{1}{c|}{}        & \multicolumn{6}{c|}{} \\ \cline{3-8}
\multicolumn{1}{c|}{}                     & \multicolumn{1}{c|}{}                       & \multicolumn{1}{c|}{\multirow{3}{*}{P}}                                                 & \multicolumn{1}{l|}{CoNLL05}                                                                     & \multicolumn{1}{l|}{3/3}    & \multicolumn{1}{l|}{3/3}   & \multicolumn{1}{l|}{3/3}       & 0/3                   \\ \cline{4-8}
\multicolumn{1}{c|}{}                     & \multicolumn{1}{c|}{}                       & \multicolumn{1}{c|}{} &                                                                    \multicolumn{1}{l|}{CoNLL09}                                                                     & \multicolumn{1}{l|}{3/3}    & \multicolumn{1}{l|}{3/3} & \multicolumn{1}{l|}{3/3}         & 3/3      \\ \cline{4-8} 
\multicolumn{1}{c|}{}                     & \multicolumn{1}{c|}{}                       & \multicolumn{1}{c|}{}                                                & \multicolumn{1}{l|}{\shortprop}                                                                     & \multicolumn{1}{l|}{3/3}    & \multicolumn{1}{l|}{\cellgreen{1/3}} & \multicolumn{1}{l|}{\cellgreen{1/3}}        &  \cellgreen{1/3}      \\ \cline{2-8} %
\end{tabular}
\end{adjustbox}
\caption{Comparing evaluation metrics on 4 examples, showing the effect of wrong predicate sense on argument label evaluation. \red{\textit{RED-italic}} shows a wrong prediction by  a hypothetical model. {\green{GREEN}} cell highlights where \shortprop\ differs from existing metrics.}
\label{tab:pred_sense}
\end{table}

%% file: cstar_table.tex
\begin{table}[]
\begin{adjustbox}{width=\linewidth}
\begin{tabular}{llllllllllll}
\multicolumn{12}{l}{\# text = Many confusing questions have been taxing my mind for years about Egypt and its people .}
\\ \hline
\multicolumn{1}{|l|}{ID}                & \multicolumn{1}{l|}{FORM}                                                                        & \multicolumn{1}{l|}{F}                  & \multicolumn{1}{l|}{\begin{tabular}[c]{@{}l@{}}PRED\\ SENSE\end{tabular}} & \multicolumn{1}{l|}{Gold} & \multicolumn{1}{l|}{P1}   & \multicolumn{1}{l|}{P2}   & \multicolumn{1}{l|}{P3}   & \multicolumn{1}{l|}{P4}   & \multicolumn{1}{l|}{P5}   & \multicolumn{1}{l|}{P6}   & \multicolumn{1}{l|}{P7}   \\ \hline
\multicolumn{1}{|l|}{1}                 & \multicolumn{1}{l|}{Many}                                                                        & \multicolumn{1}{l|}{\_}                 & \multicolumn{1}{l|}{\_}                                                   & \multicolumn{1}{l|}{\_}   & \multicolumn{1}{l|}{\_}   & \multicolumn{1}{l|}{\_}   & \multicolumn{1}{l|}{\_}   & \multicolumn{1}{l|}{\_}   & \multicolumn{1}{l|}{\_}   & \multicolumn{1}{l|}{\_}   & \multicolumn{1}{l|}{\_}   \\ \hline
\multicolumn{1}{|l|}{2}                 & \multicolumn{1}{l|}{confusing}                                                                   & \multicolumn{1}{l|}{\_}                 & \multicolumn{1}{l|}{\_}                                                   & \multicolumn{1}{l|}{\_}   & \multicolumn{1}{l|}{\_}   & \multicolumn{1}{l|}{\_}   & \multicolumn{1}{l|}{\_}   & \multicolumn{1}{l|}{\_}   & \multicolumn{1}{l|}{\_}   & \multicolumn{1}{l|}{\_}   & \multicolumn{1}{l|}{\_}   \\ \hline
\multicolumn{1}{|l|}{3}                 & \multicolumn{1}{l|}{questions}                                                                   & \multicolumn{1}{l|}{\_}                 & \multicolumn{1}{l|}{\_}                                                   & \multicolumn{1}{l|}{\parg{0}}   & \multicolumn{1}{l|}{\parg{0}}   & \multicolumn{1}{l|}{\red{\textit{\parg{1}}}}   & \multicolumn{1}{l|}{\red{\textit{\parg{1}}}}   & \multicolumn{1}{l|}{\red{\textit{\pargc{0}}}} & \multicolumn{1}{l|}{\red{\textit{\parg{1}}}}   & \multicolumn{1}{l|}{\red{\textit{\pargc{0}}}} & \multicolumn{1}{l|}{\red{\textit{\pargc{0}}}} \\ \hline
\multicolumn{1}{|l|}{4}                 & \multicolumn{1}{l|}{have}                                                                        & \multicolumn{1}{l|}{Y}                  & \multicolumn{1}{l|}{\_}                                                   & \multicolumn{1}{l|}{\_}   & \multicolumn{1}{l|}{\_}   & \multicolumn{1}{l|}{\_}   & \multicolumn{1}{l|}{\_}   & \multicolumn{1}{l|}{\_}   & \multicolumn{1}{l|}{\_}   & \multicolumn{1}{l|}{\_}   & \multicolumn{1}{l|}{\_}   \\ \hline
\multicolumn{1}{|l|}{5}                 & \multicolumn{1}{l|}{been}                                                                        & \multicolumn{1}{l|}{Y}                  & \multicolumn{1}{l|}{\_}                                                   & \multicolumn{1}{l|}{\_}   & \multicolumn{1}{l|}{\_}   & \multicolumn{1}{l|}{\_}   & \multicolumn{1}{l|}{\_}   & \multicolumn{1}{l|}{\_}   & \multicolumn{1}{l|}{\_}   & \multicolumn{1}{l|}{\_}   & \multicolumn{1}{l|}{}     \\ \hline
\multicolumn{1}{|l|}{6}                 & \multicolumn{1}{l|}{taxing}                                                                      & \multicolumn{1}{l|}{Y}                  & \multicolumn{1}{l|}{tax.01}                                               & \multicolumn{1}{l|}{\_}   & \multicolumn{1}{l|}{\_}   & \multicolumn{1}{l|}{\_}   & \multicolumn{1}{l|}{\_}   & \multicolumn{1}{l|}{\_}   & \multicolumn{1}{l|}{\_}   & \multicolumn{1}{l|}{\_}   & \multicolumn{1}{l|}{\_}   \\ \hline
\multicolumn{1}{|l|}{7}                 & \multicolumn{1}{l|}{my}                                                                          & \multicolumn{1}{l|}{\_}                 & \multicolumn{1}{l|}{\_}                                                   & \multicolumn{1}{l|}{\_}   & \multicolumn{1}{l|}{\_}   & \multicolumn{1}{l|}{\_}   & \multicolumn{1}{l|}{\_}   & \multicolumn{1}{l|}{\_}   & \multicolumn{1}{l|}{\_}   & \multicolumn{1}{l|}{\_}   & \multicolumn{1}{l|}{\_}   \\ \hline
\multicolumn{1}{|l|}{8}                 & \multicolumn{1}{l|}{mind}                                                                        & \multicolumn{1}{l|}{\_}                 & \multicolumn{1}{l|}{\_}                                                   & \multicolumn{1}{l|}{\parg{2}}   & \multicolumn{1}{l|}{\parg{2}}   & \multicolumn{1}{l|}{\parg{2}}   & \multicolumn{1}{l|}{\parg{2}}   & \multicolumn{1}{l|}{\parg{2}}   & \multicolumn{1}{l|}{\parg{2}}   & \multicolumn{1}{l|}{\parg{2}}   & \multicolumn{1}{l|}{\parg{2}}   \\ \hline
\multicolumn{1}{|l|}{9}                 & \multicolumn{1}{l|}{for}                                                                         & \multicolumn{1}{l|}{\_}                 & \multicolumn{1}{l|}{\_}                                                   & \multicolumn{1}{l|}{\_}   & \multicolumn{1}{l|}{\_}   & \multicolumn{1}{l|}{\_}   & \multicolumn{1}{l|}{\_}   & \multicolumn{1}{l|}{\_}   & \multicolumn{1}{l|}{\_}   & \multicolumn{1}{l|}{\_}   & \multicolumn{1}{l|}{\_}   \\ \hline
\multicolumn{1}{|l|}{10}                & \multicolumn{1}{l|}{years}                                                                       & \multicolumn{1}{l|}{\_}                 & \multicolumn{1}{l|}{\_}                                                   & \multicolumn{1}{l|}{\texttt{TMP}}  & \multicolumn{1}{l|}{\texttt{TMP}}  & \multicolumn{1}{l|}{\texttt{TMP}}  & \multicolumn{1}{l|}{\texttt{TMP}}  & \multicolumn{1}{l|}{\texttt{TMP}}  & \multicolumn{1}{l|}{\texttt{TMP}}  & \multicolumn{1}{l|}{\texttt{TMP}}  & \multicolumn{1}{l|}{\texttt{TMP}}  \\ \hline
\multicolumn{1}{|l|}{11}                & \multicolumn{1}{l|}{about}                                                                       & \multicolumn{1}{l|}{\_}                 & \multicolumn{1}{l|}{\_}                                                   & \multicolumn{1}{l|}{\_}   & \multicolumn{1}{l|}{\_}   & \multicolumn{1}{l|}{\_}   & \multicolumn{1}{l|}{\_}   & \multicolumn{1}{l|}{\_}   & \multicolumn{1}{l|}{\_}   & \multicolumn{1}{l|}{\_}   & \multicolumn{1}{l|}{\_}   \\ \hline
\multicolumn{1}{|l|}{12}                & \multicolumn{1}{l|}{Egypt}                                                                       & \multicolumn{1}{l|}{\_}                 & \multicolumn{1}{l|}{\_}                                                   & \multicolumn{1}{l|}{\pargc{0}} & \multicolumn{1}{l|}{\red{\textit{\pargc{1}}}} & \multicolumn{1}{l|}{\pargc{0}} & \multicolumn{1}{l|}{\red{\textit{\pargc{1}}}} & \multicolumn{1}{l|}{\red{\textit{\parg{0}}}}   & \multicolumn{1}{l|}{\red{\textit{\pargc{2}}}} & \multicolumn{1}{l|}{\pargc{0}} & \multicolumn{1}{l|}{\red{\_}}   \\ \hline
\multicolumn{1}{|l|}{13}                & \multicolumn{1}{l|}{and}                                                                         & \multicolumn{1}{l|}{\_}                 & \multicolumn{1}{l|}{\_}                                                   & \multicolumn{1}{l|}{\_}   & \multicolumn{1}{l|}{\_}   & \multicolumn{1}{l|}{\_}   & \multicolumn{1}{l|}{\_}   & \multicolumn{1}{l|}{\_}   & \multicolumn{1}{l|}{\_}   & \multicolumn{1}{l|}{\_}   & \multicolumn{1}{l|}{\_}   \\ \hline
\multicolumn{1}{|l|}{14}                & \multicolumn{1}{l|}{its}                                                                         & \multicolumn{1}{l|}{\_}                 & \multicolumn{1}{l|}{\_}                                                   & \multicolumn{1}{l|}{\_}   & \multicolumn{1}{l|}{\_}   & \multicolumn{1}{l|}{\_}   & \multicolumn{1}{l|}{\_}   & \multicolumn{1}{l|}{\_}   & \multicolumn{1}{l|}{\_}   & \multicolumn{1}{l|}{\_}   & \multicolumn{1}{l|}{\_}   \\ \hline
\multicolumn{1}{|l|}{15}                & \multicolumn{1}{l|}{people}                                                                      & \multicolumn{1}{l|}{\_}                 & \multicolumn{1}{l|}{\_}                                                   & \multicolumn{1}{l|}{\_}   & \multicolumn{1}{l|}{\_}   & \multicolumn{1}{l|}{\_}   & \multicolumn{1}{l|}{\_}   & \multicolumn{1}{l|}{\_}   & \multicolumn{1}{l|}{\_}   & \multicolumn{1}{l|}{\_}   & \multicolumn{1}{l|}{\_}   \\ \hline
\multicolumn{1}{|l|}{16}                & \multicolumn{1}{l|}{.}                                                                           & \multicolumn{1}{l|}{\_}                 & \multicolumn{1}{l|}{\_}                                                   & \multicolumn{1}{l|}{\_}   & \multicolumn{1}{l|}{\_}   & \multicolumn{1}{l|}{\_}   & \multicolumn{1}{l|}{\_}   & \multicolumn{1}{l|}{\_}   & \multicolumn{1}{l|}{\_}   & \multicolumn{1}{l|}{\_}   & \multicolumn{1}{l|}{\_}   \\ \hline
\multicolumn{12}{c}{}                                                                                                                                                                                                                                                                                                                                                                                                                                                                            \\ \cline{2-12} 
\multicolumn{1}{c|}{\multirow{11}{*}{}} & \multicolumn{1}{c|}{\multirow{5}{*}{\begin{tabular}[c]{@{}c@{}}Argument \\ HEAD \\ Evaluation\end{tabular}}} & \multicolumn{1}{c|}{\multirow{2}{*}{R}} & \multicolumn{1}{l|}{Conll09}                                              & \multicolumn{1}{l|}{4/4}  & \multicolumn{1}{l|}{3/4}  & \multicolumn{1}{l|}{3/4}  & \multicolumn{1}{l|}{2/4}  & \multicolumn{1}{l|}{2/4}  & \multicolumn{1}{l|}{2/4}  & \multicolumn{1}{l|}{3/4}  & \multicolumn{1}{l|}{2/4}  \\ \cline{4-12} 
\multicolumn{1}{c|}{}                   & \multicolumn{1}{c|}{}                                                                            & \multicolumn{1}{c|}{}                   & \multicolumn{1}{l|}{\shortprop}                                              & \multicolumn{1}{l|}{\cellgreen{3/3}}  & \multicolumn{1}{l|}{\cellgreen{2/3}}  & \multicolumn{1}{l|}{\cellgreen{2/3}}  & \multicolumn{1}{l|}{\cellgreen{2/3}}  & \multicolumn{1}{l|}{\cellgreen{3/3}}  & \multicolumn{1}{l|}{\cellgreen{1/3}}  & \multicolumn{1}{l|}{\cellgreen{3/3}}  & \multicolumn{1}{l|}{\cellgreen{2/3}}  \\ \cline{3-12} 
\multicolumn{1}{c|}{}                   & \multicolumn{1}{c|}{}                                                                            & \multicolumn{10}{c|}{}                                                                                                                                                                                                                                                                                                                              \\ \cline{3-12} 
\multicolumn{1}{c|}{}                   & \multicolumn{1}{c|}{}                                                                            & \multicolumn{1}{c|}{\multirow{2}{*}{P}} & \multicolumn{1}{l|}{Conll09}                                              & \multicolumn{1}{l|}{4/4}  & \multicolumn{1}{l|}{3/4}  & \multicolumn{1}{l|}{3/4}  & \multicolumn{1}{l|}{2/4}  & \multicolumn{1}{l|}{2/4}  & \multicolumn{1}{l|}{2/4}  & \multicolumn{1}{l|}{3/4}  & \multicolumn{1}{l|}{2/3}  \\ \cline{4-12} 
\multicolumn{1}{c|}{}                   & \multicolumn{1}{c|}{}                                                                            & \multicolumn{1}{c|}{}                   & \multicolumn{1}{l|}{\shortprop}                                              & \multicolumn{1}{l|}{\cellgreen{3/3}}  & \multicolumn{1}{l|}{\cellgreen{2/4}}  & \multicolumn{1}{l|}{\cellgreen{2/4}}  & \multicolumn{1}{l|}{\cellgreen{2/3}}  & \multicolumn{1}{l|}{\cellgreen{3/3}}  & \multicolumn{1}{l|}{\cellgreen{1/3}}  & \multicolumn{1}{l|}{\cellgreen{3/3}}  & \multicolumn{1}{l|}{\cellgreen{2/3}}  \\ \cline{2-12} 
\multicolumn{1}{c}{}                   & \multicolumn{11}{c}{}                                                                                                                                                                                                                                                                                                                                                                                                                                  \\ \cline{2-12} 
\multicolumn{1}{c|}{}                   & \multicolumn{1}{c|}{\multirow{5}{*}{\begin{tabular}[c]{@{}c@{}}Argument \\ SPAN\\  Evaluation\end{tabular}}} & \multicolumn{1}{c|}{\multirow{2}{*}{R}} & \multicolumn{1}{l|}{Conll05}                                              & \multicolumn{1}{l|}{3/3}  & \multicolumn{1}{l|}{2/3}  & \multicolumn{1}{l|}{2/3}  & \multicolumn{1}{l|}{2/3}  & \multicolumn{1}{l|}{2/3}  & \multicolumn{1}{l|}{1/3}  & \multicolumn{1}{l|}{2/3}  & \multicolumn{1}{l|}{2/3}  \\\cline{4-12} 
\multicolumn{1}{c|}{}                   & \multicolumn{1}{c|}{}                                                                            & \multicolumn{1}{c|}{}                   & \multicolumn{1}{l|}{\shortprop}                                              & \multicolumn{1}{l|}{3/3}  & \multicolumn{1}{l|}{2/3}  & \multicolumn{1}{l|}{2/3}  & \multicolumn{1}{l|}{2/3}  & \multicolumn{1}{l|}{\cellgreen{3/3}}  & \multicolumn{1}{l|}{1/3}  & \multicolumn{1}{l|}{\cellgreen{3/3}}  & \multicolumn{1}{l|}{2/3}  \\ \cline{3-12} 
\multicolumn{1}{c|}{}                   & \multicolumn{1}{c|}{}                                                                            & \multicolumn{10}{c|}{}                                                                                                                                                                                                                                                                                                                              \\ \cline{3-12} 
\multicolumn{1}{c|}{}                   & \multicolumn{1}{c|}{}                                                                            & \multicolumn{1}{c|}{\multirow{2}{*}{P}} & \multicolumn{1}{l|}{Conll05}                                              & \multicolumn{1}{l|}{3/3}  & \multicolumn{1}{l|}{2/4}  & \multicolumn{1}{l|}{2/4}  & \multicolumn{1}{l|}{2/3}  & \multicolumn{1}{l|}{2/4}  & \multicolumn{1}{l|}{1/3}  & \multicolumn{1}{l|}{2/3}  & \multicolumn{1}{l|}{2/3}  \\  \cline{4-12} 
\multicolumn{1}{c|}{}                   & \multicolumn{1}{c|}{}                                                                            & \multicolumn{1}{c|}{}                   & \multicolumn{1}{l|}{\shortprop}                                              & \multicolumn{1}{l|}{3/3}  & \multicolumn{1}{l|}{2/4}  & \multicolumn{1}{l|}{2/4}  & \multicolumn{1}{l|}{2/3}  & \multicolumn{1}{l|}{\cellgreen{3/3}}  & \multicolumn{1}{l|}{1/3}  & \multicolumn{1}{l|}{\cellgreen{3/3}}  & \multicolumn{1}{l|}{2/3}  \\ \cline{2-12} 
\end{tabular}

\end{adjustbox}
\caption{Comparing evaluation metrics on 7 examples, showing the effect of \pargcstar\ labels. 
\red{\textit{RED-italic}} and {\green{GREEN}} cell are used in the same manner as Table~\ref{tab:pred_sense}.}
\label{tab:example_c}
\end{table}

%% file: rstar_table.tex

\begin{table}[]
\begin{adjustbox}{width=\linewidth}
\begin{tabular}{lllllllllll}
\multicolumn{11}{l}{\# text = This is exactly a road that leads nowhere.}
\\ \hline
\multicolumn{1}{|l|}{ID}               & \multicolumn{1}{l|}{FORM}                                                                        & \multicolumn{1}{l|}{F}                       & \multicolumn{1}{l|}{\begin{tabular}[c]{@{}l@{}}PRED\\ SENSE\end{tabular}} & \multicolumn{1}{l|}{Gold} & \multicolumn{1}{l|}{P1}   & \multicolumn{1}{l|}{P2}   & \multicolumn{1}{l|}{P3}   & \multicolumn{1}{l|}{P4}   & \multicolumn{1}{l|}{P5}   & \multicolumn{1}{l|}{P6}   \\ \hline
\multicolumn{1}{|l|}{1}                & \multicolumn{1}{l|}{This}                                                                        & \multicolumn{1}{l|}{\_}                         & \multicolumn{1}{l|}{\_}                                                   & \multicolumn{1}{l|}{\_}   & \multicolumn{1}{l|}{\_}   & \multicolumn{1}{l|}{\_}   & \multicolumn{1}{l|}{\_}   & \multicolumn{1}{l|}{\_}   & \multicolumn{1}{l|}{\_}   & \multicolumn{1}{l|}{\_}   \\ \hline
\multicolumn{1}{|l|}{2}                & \multicolumn{1}{l|}{is}                                                                          & \multicolumn{1}{l|}{\_}                         & \multicolumn{1}{l|}{\_}                                                   & \multicolumn{1}{l|}{\_}   & \multicolumn{1}{l|}{\_}   & \multicolumn{1}{l|}{\_}   & \multicolumn{1}{l|}{\_}   & \multicolumn{1}{l|}{\_}   & \multicolumn{1}{l|}{\_}   & \multicolumn{1}{l|}{\_}   \\ \hline
\multicolumn{1}{|l|}{3}                & \multicolumn{1}{l|}{exactly}                                                                     & \multicolumn{1}{l|}{\_}                         & \multicolumn{1}{l|}{\_}                                                   & \multicolumn{1}{l|}{\_}   & \multicolumn{1}{l|}{\_}   & \multicolumn{1}{l|}{\_}   & \multicolumn{1}{l|}{\_}   & \multicolumn{1}{l|}{\_}   & \multicolumn{1}{l|}{\_}   & \multicolumn{1}{l|}{\_}   \\ \hline
\multicolumn{1}{|l|}{4}                & \multicolumn{1}{l|}{a}                                                                           & \multicolumn{1}{l|}{\_}                         & \multicolumn{1}{l|}{\_}                                                   & \multicolumn{1}{l|}{\_}   & \multicolumn{1}{l|}{\_}   & \multicolumn{1}{l|}{\_}   & \multicolumn{1}{l|}{\_}   & \multicolumn{1}{l|}{\red{\textit{\parg{0}}}}   & \multicolumn{1}{l|}{\_}   & \multicolumn{1}{l|}{\_}   \\ \hline
\multicolumn{1}{|l|}{5}                & \multicolumn{1}{l|}{road}                                                                        & \multicolumn{1}{l|}{\_}                         & \multicolumn{1}{l|}{\_}                                                   & \multicolumn{1}{l|}{\parg{0}}   & \multicolumn{1}{l|}{\red{\textit{\parg{1}}}}   & \multicolumn{1}{l|}{\parg{0}}   & \multicolumn{1}{l|}{\red{\textit{\parg{1}}}}   & \multicolumn{1}{l|}{\red{\textit{\_}}}   & \multicolumn{1}{l|}{\red{\textit{\pargr{0}}}} & \multicolumn{1}{l|}{\red{\textit{\pargr{0}}}} \\ \hline
\multicolumn{1}{|l|}{6}                & \multicolumn{1}{l|}{that}                                                                        & \multicolumn{1}{l|}{\_}                         & \multicolumn{1}{l|}{\_}                                                   & \multicolumn{1}{l|}{\pargr{0}} & \multicolumn{1}{l|}{\pargr{0}} & \multicolumn{1}{l|}{\red{\textit{\pargr{1}}}} & \multicolumn{1}{l|}{\red{\textit{\pargr{1}}}} & \multicolumn{1}{l|}{\pargr{0}} & \multicolumn{1}{l|}{\pargr{0}} & \multicolumn{1}{l|}{\red{\textit{\parg{0}}}}   \\ \hline
\multicolumn{1}{|l|}{7}                & \multicolumn{1}{l|}{leads}                                                                       & \multicolumn{1}{l|}{Y}                          & \multicolumn{1}{l|}{lead.01}                                              & \multicolumn{1}{l|}{\_}   & \multicolumn{1}{l|}{\_}   & \multicolumn{1}{l|}{\_}   & \multicolumn{1}{l|}{\_}   & \multicolumn{1}{l|}{\_}   & \multicolumn{1}{l|}{\_}   & \multicolumn{1}{l|}{\_}   \\ \hline
\multicolumn{1}{|l|}{8}                & \multicolumn{1}{l|}{nowhere}                                                                     & \multicolumn{1}{l|}{\_}                         & \multicolumn{1}{l|}{\_}                                                   & \multicolumn{1}{l|}{\parg{4}}   & \multicolumn{1}{l|}{\parg{4}}   & \multicolumn{1}{l|}{\parg{4}}   & \multicolumn{1}{l|}{\parg{4}}   & \multicolumn{1}{l|}{\parg{4}}   & \multicolumn{1}{l|}{\parg{4}}   & \multicolumn{1}{l|}{\parg{4}}   \\ \hline
\multicolumn{1}{|l|}{9}                & \multicolumn{1}{l|}{.}                                                                           & \multicolumn{1}{l|}{\_}                         & \multicolumn{1}{l|}{\_}                                                   & \multicolumn{1}{l|}{\_}   & \multicolumn{1}{l|}{\_}   & \multicolumn{1}{l|}{\_}   & \multicolumn{1}{l|}{\_}   & \multicolumn{1}{l|}{\_}   & \multicolumn{1}{l|}{\_}   & \multicolumn{1}{l|}{\_}   \\ \hline
\multicolumn{1}{c}{\multirow{10}{*}{}} & \multicolumn{10}{c}{} \\ 
\cline{2-11} 
\multicolumn{1}{c}{}                   & \multicolumn{1}{|c|}{\multirow{5}{*}{\begin{tabular}[c]{@{}c@{}}Argument \\ HEAD \\ Evaluation\end{tabular}}} & \multicolumn{1}{c|}{\multirow{2}{*}{R}}    & \multicolumn{1}{l|}{Conll09}                                              & \multicolumn{1}{l|}{3/3}  & \multicolumn{1}{l|}{2/3}  & \multicolumn{1}{l|}{2/3}  & \multicolumn{1}{l|}{1/3}  & \multicolumn{1}{l|}{2/3}  & \multicolumn{1}{l|}{2/3}  & \multicolumn{1}{l|}{1/3}  \\ \cline{4-11} 
\multicolumn{1}{c}{}                   & \multicolumn{1}{|c|}{}                                                                            & \multicolumn{1}{c|}{}                           & \multicolumn{1}{l|}{\shortprop}                                             & \multicolumn{1}{l|}{3/3}  & \multicolumn{1}{l|}{\cellgreen{1/3}}  & \multicolumn{1}{l|}{2/3}  & \multicolumn{1}{l|}{1/3}  & \multicolumn{1}{l|}{\cellgreen{1/3}}  & \multicolumn{1}{l|}{\cellgreen{1/3}}  & \multicolumn{1}{l|}{1/3}  \\ \cline{3-11} 
\multicolumn{1}{c|}{}        & \multicolumn{1}{c|}{}        & \multicolumn{9}{c|}{} \\ \cline{3-11}

\multicolumn{1}{c}{}                   & \multicolumn{1}{|c|}{}                                                                            & \multicolumn{1}{c|}{\multirow{2}{*}{P}} & \multicolumn{1}{l|}{Conll09}                                              & \multicolumn{1}{l|}{3/3}  & \multicolumn{1}{l|}{2/3}  & \multicolumn{1}{l|}{2/3}  & \multicolumn{1}{l|}{1/3}  & \multicolumn{1}{l|}{2/3}  & \multicolumn{1}{l|}{2/3}  & \multicolumn{1}{l|}{1/3}  \\ \cline{4-11} 
\multicolumn{1}{c}{}                   & \multicolumn{1}{|c|}{}                                                                            & \multicolumn{1}{c|}{}                           & \multicolumn{1}{l|}{\shortprop}                                             & \multicolumn{1}{l|}{3/3}  & \multicolumn{1}{l|}{\cellgreen{1/3}}  & \multicolumn{1}{l|}{2/3}  & \multicolumn{1}{l|}{1/3}  & \multicolumn{1}{l|}{\cellgreen{1/3}}  & \multicolumn{1}{l|}{\cellgreen{1/3}}  & \multicolumn{1}{l|}{1/3}  \\ \cline{2-11} 
\multicolumn{1}{c}{}                   & \multicolumn{10}{c}{}\\ 
\cline{2-11} 
\multicolumn{1}{c}{}                   & \multicolumn{1}{|c|}{\multirow{5}{*}{\begin{tabular}[c]{@{}c@{}}Argument \\ SPAN\\  Evaluation\end{tabular}}} & \multicolumn{1}{c|}{\multirow{2}{*}{R}}    & \multicolumn{1}{l|}{Conll05}                                              & \multicolumn{1}{l|}{3/3}  & \multicolumn{1}{l|}{2/3}  & \multicolumn{1}{l|}{2/3}  & \multicolumn{1}{l|}{1/3}  & \multicolumn{1}{l|}{2/3}  & \multicolumn{1}{l|}{1/3}  & \multicolumn{1}{l|}{1/3}  \\ \cline{4-11} 
\multicolumn{1}{c}{}                   & \multicolumn{1}{|c|}{}                                                                            & \multicolumn{1}{c|}{}                           & \multicolumn{1}{l|}{\shortprop}                                             & \multicolumn{1}{l|}{3/3}  & \multicolumn{1}{l|}{\cellgreen{1/3}}  & \multicolumn{1}{l|}{2/3}  & \multicolumn{1}{l|}{1/3}  & \multicolumn{1}{l|}{\cellgreen{1/3}}  & \multicolumn{1}{l|}{1/3}  & \multicolumn{1}{l|}{1/3}  \\ \cline{3-11} 
\multicolumn{1}{c|}{}        & \multicolumn{1}{c|}{}        & \multicolumn{9}{c|}{} \\ \cline{3-11}
\multicolumn{1}{c}{}                   & \multicolumn{1}{|c|}{}                                                                            & \multicolumn{1}{c|}{\multirow{2}{*}{P}} & \multicolumn{1}{l|}{Conll05}                                              & \multicolumn{1}{l|}{3/3}  & \multicolumn{1}{l|}{2/3}  & \multicolumn{1}{l|}{2/3}  & \multicolumn{1}{l|}{1/3}  & \multicolumn{1}{l|}{2/3}  & \multicolumn{1}{l|}{1/3}  & \multicolumn{1}{l|}{1/3}  \\ \cline{4-11} 
\multicolumn{1}{c}{}                   & \multicolumn{1}{|c|}{}                                                                            & \multicolumn{1}{c|}{}                           & \multicolumn{1}{l|}{\shortprop}                                             & \multicolumn{1}{l|}{3/3}  & \multicolumn{1}{l|}{\cellgreen{1/3}}  & \multicolumn{1}{l|}{2/3}  & \multicolumn{1}{l|}{1/3}  & \multicolumn{1}{l|}{\cellgreen{1/3}}  & \multicolumn{1}{l|}{1/3}  & \multicolumn{1}{l|}{1/3}  \\ \cline{2-11} 
\end{tabular}
\end{adjustbox}
\caption{Comparing evaluation metrics on 6 examples, showing the effect of \pargrstar\ labels. 
\red{\textit{RED-italic}} and {\green{GREEN}} cell are used in the same manner as Table~\ref{tab:pred_sense}.}
\label{tab:example_r}
\end{table}

%% file: compare_table.tex
\begin{table*}[tbhp]
\centering
\begin{adjustbox}{width=\linewidth}
\begin{tabular}{ccllllllllll}
\toprule
\multirow{3}{*}{Model}        & \multirow{3}{*}{\begin{tabular}[c]{@{}c@{}}Evaluation\\ script\end{tabular}}   & \multicolumn{5}{c}{In-domain}                       & \multicolumn{5}{c}{Out-of-domain}                    \\ \cmidrule(lr){3-7} \cmidrule(lr){8-12}
                              &                              & \multicolumn{1}{c}{PSD}   & \multicolumn{3}{c}{Argument Classification} &    & \multicolumn{1}{c}{PSD}    & \multicolumn{3}{c}{Argument Classification}     & \\  \cmidrule(lr){3-3} \cmidrule(lr){4-6} \cmidrule(lr){8-8} \cmidrule(lr){9-11}
                              &                              & \multicolumn{1}{c}{F1}    & P             & R            & \multicolumn{1}{c}{F1}    & \multicolumn{1}{c}{(r)}       & \multicolumn{1}{c}{F1}     & P             & R            & \multicolumn{1}{c}{F1}   & \multicolumn{1}{c}{(r)}        \\ \midrule
                              
\multirow{2}{*}{\cite{conia-etal-2021-unifying}} & \multicolumn{1}{l}{CoNLL09}  & 96.9  & 89.5          & 89.5         & 89.5    & (3)    & 87.8   & 82.0          & 81.9         & 81.9 & (3)        \\
& \shortprop & 95.5($\downarrow$\red{1.4})  & 86.6          & 86.6         & 86.6($\downarrow$\red{2.9}) & (2)        & 80.9($\downarrow$\red{6.9})   & 72.4         & 72.6         & 72.5($\downarrow$\red{9.4})  & (4)  \\  \cmidrule{2-12}

\multirow{2}{*}{\cite{blloshmi2021generating}$_{\text{nested}}$} & \multicolumn{1}{l}{CoNLL09}  & 97.1  & 89.3          & 81.9         & 85.4 & (4)       & 89.7   & 82.8          & 75.7         & 79.1  & (4)       \\ 
& \shortprop & 96.4($\downarrow$\red{0.7}) & 86.8 & 79.8 & 83.1($\downarrow$\red{2.3}) & (4) & 86.7($\downarrow$\red{3.0}) & 75.7 & 69.9 & 72.7($\downarrow$\red{6.4}) & (3)  \\ \cmidrule{2-12}

\multirow{2}{*}{\cite{blloshmi2021generating}$_{\text{flat}}$} & \multicolumn{1}{l}{CoNLL09}  & 97.4  &  90.9         &  89.6        & \textbf{90.2}  &   (1)    & 90.1   & 83.9          & 82.1         & 83.0 & (2)        \\ 
& \shortprop &  96.9($\downarrow$\red{0.5}) &     88.6     &     87.4    &   \underline{\textbf{88.0}}($\downarrow$\red{2.2}) &   (1)   & 87.8($\downarrow$\red{2.3})   & 77.6        & 76.3         & \underline{\textbf{76.9}}($\downarrow$\red{6.1}) & (1)  \\ \cmidrule{2-12}

\multirow{2}{*}{\cite{jindal-EtAl:2022:LREC}} & \multicolumn{1}{l}{CoNLL09}  & 96.8  & 89.9          & 89.3         & 89.6 & (2)        & 89.8   & 82.9          & 83.1         & \textbf{83.02} & (1)        \\
& \shortprop &
                              95.5($\downarrow$\red{1.3})  & 86.8          & 86.3         & 86.55($\downarrow$\red{3.0})    & (3)      & 83.4($\downarrow$\red{6.4})   & 73.9          & 74.3         & 74.1($\downarrow$\red{8.9}) &(2)   \\
                              \bottomrule
\end{tabular}
\end{adjustbox}
\caption{Comparison of SoTA SRL models with \shortprop\ and CoNLL09 evaluation metrics on CoNLL09 dataset. (r) denotes the ranking of SRL models corresponding to the evaluation metric. \textbf{BOLD} shows the best model with CoNLL09 evaluation script and \underline{\textbf{BOLD}} shows the best SRL model with \shortprop.}
\label{tab:p_09}
\end{table*}

\begin{table}[tbhp]
\centering
\begin{adjustbox}{width=\linewidth}
\begin{tabular}{ccllll}
\toprule
\multicolumn{1}{c}{Dataset}                    & \multicolumn{1}{l}{Args} & \multicolumn{1}{l}{Train} & \multicolumn{1}{l}{Dev}  & \multicolumn{1}{l}{Test} & ood  \\ \midrule
\multicolumn{1}{c}{\multirow{2}{*}{CoNLL09}}   & \multicolumn{1}{l}{\pargcstar}   & \multicolumn{1}{l}{0.77}  & \multicolumn{1}{l}{1.05} & \multicolumn{1}{l}{0.88} & 1.15 \\ \cmidrule{2-6} 
\multicolumn{1}{c}{}                           & \multicolumn{1}{l}{\pargrstar}   & \multicolumn{1}{l}{1.98}  & \multicolumn{1}{l}{2.03} & \multicolumn{1}{l}{2.07} & 2.24 \\ \midrule
\multicolumn{1}{c}{\multirow{2}{*}{CoNLL05}}   & \multicolumn{1}{l}{\pargcstar}   & \multicolumn{1}{l}{1.22}  & \multicolumn{1}{l}{1.24} & \multicolumn{1}{l}{1.71} & 0.91 \\ \cmidrule{2-6} 
\multicolumn{1}{c}{}                           & \multicolumn{1}{l}{\pargrstar}   & \multicolumn{1}{l}{3.26}  & \multicolumn{1}{l}{3.36} & \multicolumn{1}{l}{3.38} & 2.91 \\ \bottomrule
\end{tabular}
\end{adjustbox}
\caption{Representation of \pargcstar\ and \pargrstar\ arguments in each split of different SRL datasets.}
\label{tab:dataset}
\end{table}

%% file: compare_table_05.tex
\begin{table*}[tbhp]
\centering
\begin{adjustbox}{width=0.9\linewidth}
\begin{tabular}{ccllllllllll}
\toprule
\multirow{3}{*}{Model}        & \multirow{3}{*}{\begin{tabular}[c]{@{}c@{}}Evaluation\\ script\end{tabular}}   & \multicolumn{5}{c}{In-domain}                       & \multicolumn{5}{c}{Out-of-domain}                    \\ \cmidrule(lr){3-7} \cmidrule(lr){8-12}
                              &                              & \multicolumn{1}{c}{PSD}   & \multicolumn{3}{c}{Argument Classification} &    & \multicolumn{1}{c}{PSD}    & \multicolumn{3}{c}{Argument Classification}     & \\  \cmidrule(lr){3-3} \cmidrule(lr){4-6} \cmidrule(lr){8-8} \cmidrule(lr){9-11}
                              &                              & \multicolumn{1}{c}{F1}    & P             & R            & \multicolumn{1}{c}{F1}    & \multicolumn{1}{c}{(r)}       & \multicolumn{1}{c}{F1}     & P             & R            & \multicolumn{1}{c}{F1}   & \multicolumn{1}{c}{(r)}        \\ \midrule
\multirow{2}{*}{\cite{zhang2021semantic}$_{\text{crf}}$} & \multicolumn{1}{l}{CoNLL05}  & 100   & 86.5          & 88.3           & 87.4 & (3)         & 100    & 79.0          & 81.1         & 80.0    & (3)     \\
                              & \multicolumn{1}{l}{\shortprop} & 100   & 86.1          & 87.8        & 87.03($\downarrow$\red{0.4}) & (2)           & 100    & 78.7          & 80.8         & 79.7($\downarrow$\red{0.3}) & (2)     \\  \cmidrule{2-12}
\multirow{2}{*}{\cite{zhang2021semantic}$_{\text{crf2o}}$} & \multicolumn{1}{l}{CoNLL05}  & 100   & 86.9          & 88.6           & 87.7 & (2)         & 100    & 78.9          & 81.2         &80.03    & (2)     \\
                              & \multicolumn{1}{l}{\shortprop} & 100   & 86.5          & 88.1         & \underline{\textbf{87.3}}($\downarrow$\red{0.4})& (1)           & 100    & 78.5          & 80.8         & 79.6($\downarrow$\red{0.4}) & (3)   \\   \cmidrule{2-12}
\multirow{2}{*}{\cite{jindal-EtAl:2022:LREC}} & \multicolumn{1}{l}{CoNLL05}  & 100   & 87.4          & 88.0           & \textbf{87.74} & (1)        & 100    & 80.4          & 81.4         & \textbf{80.9}  & (1)       \\
                              & \multicolumn{1}{l}{\shortprop} & 100   & 86.8          & 87.1         & 87.0($\downarrow$\red{0.7}) & (3)          & 100    & 79.7          & 80.5         & \underline{\textbf{80.1}}($\downarrow$\red{0.8})& (1)    \\  
                              \bottomrule
\end{tabular}
\end{adjustbox}
\caption{Comparison of SoTA SRL models with \shortprop\ and CoNLL05 evaluation metrics on CoNLL05 dataset. (r) denotes the ranking of SRL models corresponding to the evaluation metric. \textbf{BOLD} shows the best model with CoNLL05 evaluation script and \underline{\textbf{BOLD}} shows the best SRL model with \shortprop.}
\label{tab:p_05}
\end{table*}

%% file: appendix.tex
\section{Some historical background to existing evaluation metrics for SRL}\label{sec:eval-metric}

Shared tasks have boosted the development of systems capable of extracting predicates and arguments from natural language sentences. Two  regular academic events have promoted SRL shared tasks: SemEval and CoNLL. In this section, we summarize the approaches to SRL evaluation in the shared tasks and categorize their shortcomings. 

\subsection{Senseval and SemEval}

SemEval (Semantic Evaluation) is a series of evaluations of computational semantic analysis systems that evolved from the Senseval (word sense evaluation) series. The SENSEVAL-3 \cite{litkowski-2004-senseval} was about the automatic labeling of semantic roles and was designed to encourage research into and use of the FrameNet dataset. The systems receive as input unsegmented sentences (the constituents are not identified) a target word and its frame. They have to identify the frame elements within that sentence and tag them with the appropriate frame element name. In general, FrameNet frames contain many frame elements (an average of 10), most of which are not instantiated in a given sentence. Systems were not penalized if they returned more frame elements than those identified in the gold data. In scoring, each frame element returned by a system was counted as an item attempted. If the frame element had been identified in the gold data, the answer was scored as correct. In addition, the scoring program required that the frame boundaries identified by the system's answer overlap with the gold annotation. An additional measure of system performance was the degree of overlap. If a system's answer coincided precisely with the start and end position in the gold data, the system received an overlap score of 1.0. If not, the overlap score was the number of characters overlapping divided by the length of the gold annotation. The number attempted was the number of non-null frame elements generated by a system. Precision was computed as the number of correct answers divided by the number attempted. The recall was computed as the number of correct answers divided by the number of frame elements in the test set. Overlap was the average overlap of all correct answers. The percent Attempted was the number of frame elements generated divided by the number of frame elements in the test set, multiplied by 100. 

At SemEval-2007, three tasks evaluate SRL. In task 17 (subtask 2), the goal of the systems was to locate the constituents, which are the arguments of a given verb, and assign them appropriate semantic roles. Systems have to annotate the corpus using two different role label sets: the PropBank and the VerbNet. SemLink mapping \cite{loper2007combining} was used to generate the VerbNet roles. The precision, recall, and F-measure for both role label sets were calculated for each system output using the \verb|srl-eval.pl| script from the CoNLL-2005 scoring package \cite{carreras-marquez-2005-introduction} (see below). Task 18 focused on Arabic and also used the same CoNLL-2005 scoring package. 
Task 19 consists of recognizing words and phrases that evoke semantic frames from FrameNet and their semantic dependents, which are usually, but not always, their syntactic dependents. The evaluation measured precision and recall for frames and frame elements, with partial credit for incorrect but closely related frames. Two types of evaluation were carried out. The first is the label matching evaluation. The participant's labeled data were compared directly with the gold standard labeled using the same evaluation procedure used in the previous SRL tasks at SemEval. The second is the semantic dependency evaluation, in which both the gold standard and the submitted data were first converted to semantic dependency graphs and compared. 

SemEval-2012 and SemEval-2013 introduced the `Spatial Role Labeling' task. It concerns the identification of trajectors, landmarks, spatial indicators, the links between them, and the type of spatial relationships, including region, direction, and distance. Although similar to the standard SRL task, we will not discuss Spatial Role Labeling and its evaluation in this paper. Starting from SemEval-2014, a deeper semantic representation of sentences in a single graph-based structure via semantic parsing substituted the `shallow' SRL tasks.

\subsection{CoNLL}

The Conference on Computational Natural Language Learning (CoNLL) is a yearly conference organized by the ACL's Special Interest Group on Natural Language Learning (SIGNLL), focusing on theoretically, cognitively and scientifically motivated approaches to computational linguistics since 1999. The 2004 and 2005 shared tasks of the CoNLL were dedicated to SRL monolingual setting (English). The CoNLL-2008 shared task proposes a unified dependency-based formalism, which models both syntactic dependencies and semantic roles. The CoNLL-2009 builds on the CoNLL-2008 task and extends it to multiple languages. 

The CoNLL-2004 shared task \cite{carreras-marquez-2004-introduction} was based on the PropBank corpus, six sections of the Wall Street Journal part of the Penn Treebank \cite{kingsbury-palmer-2002-treebank} enriched with predicate–argument structures. 
The participants need to come up with machine learning strategies to SRL on the basis of only partial syntactic information, avoiding the use of full parsers and external lexico-semantic knowledge bases. The annotations provided for the development of systems include, apart from the argument boundaries and role labels, the levels of processing treated in the previous editions of the CoNLL shared task, i.e., words, PoS tags, base chunks, clauses, and named entities. In practice, number of target verbs are marked in a sentence, each governing one proposition. A system has to recognize and label the arguments of each target verb. The systems were evaluated with respect to precision, recall and the F1 measure using the \verb|srl-eval.pl| program. For an argument to be correctly recognized, the words spanning the argument as well as its semantic role have to be correct. The verb argument is the lexicalization of the predicate of the proposition. Most of the time, the verb corresponds to the target verb of the proposition, which is provided as input, and only in few cases the verb participant spans more words than the target verb. This situation makes the verb easy to identify and, since there is one verb with each proposition, evaluating its recognition overestimates the overall performance of a system. For this reason, the verb argument is excluded from evaluation. The shared task proceedings does not details how non-continuous arguments are evaluated.

Compared to the shared task of CoNLL-2004, three novelties were introduced in the 2005 edition \cite{carreras2005introduction}: 1) the complete syntactic trees, with information of the lexical head for each syntactic constituent, given by two alternative parsers have been provided as input; 2) the training corpus has been substantially enlarged; 3) a cross-corpora evaluation is performed using a fresh test set from the Brown corpus. Evaluation didn't changed compared to CoNLL-2004 and it was reported to use the same evaluation code, a system has to recognize and label the arguments of each target verb. To support the role labeling task, sentences contain input annotations, that consist of syntactic information and named entities. Evaluation is performed on a collection of unseen test sentences, that are marked with target verbs and contain only predicted input annotations.

The CoNLL 2008 shared task \cite{surdeanu-etal-2008-conll} was dedicated to the joint parsing of syntactic and semantic dependencies. The shared task was divided into three subtasks: (i) parsing of syntactic dependencies, (ii) identification and disambiguation of semantic predicates, and (iii) identification of arguments and assignment of semantic roles for each predicate. SRL was performed and evaluated using a dependency-based representation for both syntactic and semantic dependencies. 

The task addressed propositions centered around both verbal and nominal predicates. The data was composed by the Penn Treebank, BBN’s named entity corpus, PropBank and NomBank. The dependency-annotated data was obtain from a conversion algorithm from the constituent analyses.  convert the underlying constituent analysis of PropBank and NomBank into a dependency analysis, the head of a semantic argument was identified with a straightforward heuristic. But there are cases that require
special treatment, some arguments ended up with several syntactic heads and some arguments that were initially discontinuous in PropBank or NomBank where merged.

The official evaluation measures consist of three different scores: (i) syntactic dependencies are scored using the labeled attachment score (LAS),
(ii) semantic dependencies are evaluated using a labeled F1 score, and (iii) the overall task is scored with a macro average of the two previous scores. The semantic propositions are evaluated by converting them to semantic dependencies, i.e., a semantic dependency from every predicate
to all its individual arguments were created. These dependencies are labeled with the labels of the corresponding arguments. Additionally, a semantic dependency from each predicate to a virtual
ROOT node was created. The latter dependencies are labeled with the predicate senses. This approach guarantees that the semantic dependency structure conceptually forms a single-rooted, connected (not necessarily acyclic) graph. More importantly, this scoring strategy implies that if a system assigns the incorrect predicate sense, it still receives some points for the arguments correctly assigned. Several additional evaluation measures were applied to further analyze the performance of the participating systems. The \emph{Exact Match} reports the percentage of sentences that are completely correct, i.e., all the generated syntactic dependencies are correct and all the semantic propositions are present and correct. The \emph{Perfect Proposition F1} score entire semantic frames or propositions. The ratio between labeled F1 score for semantic dependencies and the LAS for syntactic dependencies.

As in CoNLL-2008, the CoNLL-2009 shared task \cite{hajic-etal-2009-conll} combined syntactic dependency parsing and the task of identifying and labeling semantic arguments of verbs or nouns for six more languages (Catalan, Chinese, Czech, German, Japanese and Spanish) in addition to the original English from CoNLL-2008. Participants can choose the joint task (syntactic dependency parsing and SRL), or SRL-only (syntactic dependency provided). The novelty is that the evaluation data indicated which words were to be dealt with (for the SRL task). Predicate disambiguation was still part of the task, whereas the identification of argument-bearing words was not. This decision was made to compensate for the significant differences between languages and between the annotation schemes used. The evaluation of SRL was done similar to CoNLL-2008.